\documentclass{article}
\pdfoutput=1

\usepackage{natbib}
\usepackage{graphicx}
\usepackage{bm}% bold math

\newcommand{\latexOrPdflatex}[2]{\ifx\undefined\pdfoutput%
#1%
\else%
#2%
\fi}

\latexOrPdflatex{
\newcommand{\href}[2]{#2}
}{
\usepackage[colorlinks,pagebackref,citecolor=blue,bookmarks=true,pdfstartview=FitH,pdfstartpage=1,pdftitle={Driving Forces and SFA},pdfauthor={Wolfgang Konen},a4paper]{hyperref}
}

\renewcommand{\vec}[1]{\bm{#1}}	% vector boldface
\newcommand{\is}{x}			% input signal
\newcommand{\tfc}{h}			% transfer function component
\newcommand{\tf}{g}			% transfer function
\newcommand{\os}{y}			% output signal

\newcommand{\uu}{w}			% time series
\newcommand{\p}{\gamma}			% true driving force
\newcommand{\pest}{\gamma^{est}}	% estimated driving force
\newcommand{\pslow}{\gamma_S}			% true driving force
\newcommand{\pfast}{\gamma_F}			% true driving force

\newcommand{\emb}{x}			% embedding vector
\newcommand{\eqref}[1]{Eq.~(\ref{#1})}
\newcommand{\Secref}[1]{Sec.~\ref{#1}}
\newcommand{\captionEM}[1]{\caption{\em #1}}

\title{%
\vspace{-0.5cm}{\small \tt \raggedleft{e-print published at
\href{http://arxiv.org/abs/0911.4397v1}{http://arxiv.org/abs/0911.4397v1} November 2009}} \\
\vspace{1.cm} 
How slow is slow? \\
SFA detects signals that are slower than the driving force
}
\author{Wolfgang Konen, Patrick Koch \\ \\
 Institute for Informatics, Cologne University of Applied Sciences \\
 Steinm\"ullerallee 1, D-51643 Gummersbach, Germany \\
 \texttt{\href{http://www.gm.fh-koeln.de/~konen}{http://www.gm.fh-koeln.de/{\footnotesize$\sim$}konen}} \\
 \texttt{\href{mailto:{wolfgang.konen,patrick.koch}@fh-koeln.de}{wolfgang.konen@fh-koeln.de}}
}

\date{}

\begin{document} 

\maketitle

\begin{abstract}
Slow feature analysis (SFA) is a method for extracting slowly varying driving forces from quickly varying nonstationary time series. We show here that it is possible for SFA to detect a component which is even slower than the driving force itself (e.g. the envelope of a modulated sine wave). It is shown that it depends on circumstances like the embedding dimension,  the time series predictability, or the base frequency, whether the driving force itself or a slower subcomponent is detected. We observe a phase transition from one regime to the other and it is the purpose of this work to quantify the influence of various parameters on this phase transition. We conclude that  {\em what} is percieved as slow by SFA varies and that a more or less fast switching from one regime to the other occurs, perhaps showing some similarity to human perception.
\end{abstract}

%PRL \keywords{driving force, nonlinear time series analysis, nonstationary time series, slow feature analysis}

\section{Introduction} \label{sec:introduction}

The analysis of nonstationary time series plays an important role in the data understanding of various phenomena such as temperature drift in experimental setup, global warming in climate data or varying heart rate in cardiology. Such nonstationarities can be modeled by underlying parameters, referred to as
 driving forces, that change the dynamics of the system smoothly on a slow
 time scale or abruptly but rarely, e.g.\ if the dynamics switches between
 different discrete states.~\cite{Wis2003c}.  
 
 Often, e.g. in EEG-analysis or in monitoring of complex chemical or electrical power plants, one is particularly interested in revealing the driving forces themselves from the raw observed time series since they show interesting aspects of the underlying dynamics, for example the switching between different dynamic regimes.  
 
Several methods for detecting and visualizing driving forces have been developed: based on recurrence
plots~\cite[]{EckmOlifRuel1987,Casd1997}, error dissimilarity matrix~\cite{Schreib1999}, feedforward ANNs with extra input unit~\cite[]{VerdGranNavo+2001} or, as Wiskott recently proposed, by Slow Feature Analysis (SFA)~\cite{Wis2003c}.
 
What is "slow" in the driving forces compared to the raw observed time series? Often it might be the case that driving forces contain components on different time scales and it is crucial to understand which time scale will be selected by the driving force algorithm. As an example we will consider driving forces made up of two overlayed frequencies $f_1  < f_2$. Will the driving force detection algorithm detect the slower one of the frequencies, $f_1$, thus being more slow, or the combined driving force made up of $f_1$ and $f_2$, thus being more accurate? With this paper we try to deepen our understanding which paramaters influence whether the first or the second choice is taken. 
 
We base our analysis on SFA~\cite{WisSej2002,Wis2003c} as driving force detection algrithm since it constitutes a versatile, robust and fast algorithm.

\section{Slow Feature Analysis}

 Slow feature analysis (SFA) has been originally developed in context of an
 abstract model of unsupervised learning of invariances in the visual
 system of vertebrates~\cite[]{Wis98a} and is described in detail
 in~\cite[]{WisSej2002,Wis2003c}.

\subsection{The SFA algorithm}
 We briefly review here the approach described in~\cite{Wis2003c}.
 The general objective of SFA is to extract slowly varying features from a
 quickly varying multidimensional signal.  For a scalar output signal and an $N$-dimensional input
 signal $\vec{\is} = \vec{\is}(t)$ where $t$ indicates time and $\vec{\is} = [\is_1,...,\is_N]^T$ is a
 vector, the question can be formalized
 as follows:  Find the input-output function $\tf(\vec{\is})$ that generates a
 scalar output signal
\begin{equation}
 \os(t) := \tf(\vec{\is}(t))
\end{equation}
 with its temporal variation as  slowly as possible, measured by the variance
 of the time derivative:
\begin{equation}
 \mbox{minimize} \quad \Delta(y) = \langle\dot{\os}^2\rangle \label{eq:slowness}
\end{equation}
 with $\langle \cdot \rangle$ indicating the temporal mean.  To avoid the trivial constant solution the output signal  has to meet the following constraints:
%
%PRL \begin{subequations} \label{eq:constr0-constr1}
\begin{eqnarray}
 \langle\os\rangle &=& 0 \quad \mbox{(zero mean)} \,, \label{eq:constr0} \\
 \langle\os^2\rangle &=& 1 \quad \mbox{(unit variance)} \label{eq:constr1} \,.
\end{eqnarray}
%PRL \end{subequations}

 This is an optimization problem of variational calculus and as such
 difficult to solve. But if we constrain the input-output function to
 be a linear combination of some fixed and possibly nonlinear basis
 functions, the problem becomes tractable with the mathematical details given in~\cite{Wis2003c}. (A typical choice for the nonlinear basis functions are monomials of degree 2, 
$$
	 \vec{\tfc}(\vec{\is}) = [\is_1,\, \is_2,\, ...,\, \is_N,\,
 												  	\is_1^2,\, \is_1 \is_2,\, ...,\, \is_N^2]^T,
$$ 																
 but other choices, e.~g. monomials of higher degree or radial basis functions could be used as well.)  Basically, SFA consists of the following four steps: (i)~expand the input
 signal with some set of fixed possibly nonlinear functions; (ii)~sphere
 the expanded signal to obtain components with zero mean and unit
 covariance matrix; (iii)~compute the time derivative of the sphered
 expanded signal and determine the normalized eigenvector of its covariance
 matrix with the smallest eigenvalue; (iv)~project the sphered expanded
 signal onto this eigenvector to obtain the output signal, which we will denote by $y_1$.
 Sometimes we will be also interested in the second-smallest eigenvalue and the corresponding projected output signal, which we will denote by $y_2$. As usual for eigenvectors, higher components are orthogonal to lower ones, i.~e. signal $\os_i$ satisfies the $i-1$ additional conditions  $\langle\os_i \os_j \rangle = 0 \quad \forall j=1,\ldots,i-1$.

 In the rest of this paper we will work with time series $\vec{\is}(1),\vec{\is}(2),\ldots,\vec{\is}(t),\, t\in \textbf{N}$, instead of
 continuous signals $\vec{\is}(t),\, t\in \textbf{R}$, but the transfer of the algorithm
 described above to time series is straight forward.  The time derivative
 is simply computed as the difference between successive data points
 assuming a constant sampling spacing $\Delta t = 1$.

\subsection{Slowness indicator $\eta$}
It is useful to have a certain measure for the slowness of a signal. In principle $\Delta(y)$ is such an indicator, but Wiskott and Sejnowski~\cite{WisSej2002} propose a more intuitive measure $\eta$ defined by
\begin{equation}
	\eta(y) = \frac{T}{2\pi}\frac{\sqrt{\Delta(y)}}
				    									 {\sqrt{\left\langle (y-\left\langle y \right\rangle)^2 \right\rangle}}
	\label{eq:eta}
\end{equation}
where $\langle \cdot \rangle$ indicates the temporal average over $[t_0,t_0+T]$. The measurement interval $T$ should be at least as long as one or two periods of the slowest signal component  It is easy to show~\cite{WisSej2002} that for a pure sine wave $y(t)=\sin(n 2\pi t/T)$ the $\eta$-value counts the number of oscillations in the observation interval, i.e. $\eta(y)=n$. The denominator in \eqref{eq:eta} ensures that each linear transform $ay(t)+b$ has the same $\eta$ as $y(t)$. (This normalization is not necessary for the training output signals of SFA, which are already by construction normalized to zero mean and unit variance, but it allows to apply the measure $\eta$ to unnormalized time series like driving forces as well. Also note that SFA output signals derived from {\em test data} are only approximately normalized, so that the normalization is useful to make $\eta$ independent of an accidental scaling factor.) Low $\eta$-values indicate slow signals, high $\eta$-values fast signals.

\section{Experiments}

 In the following we present examples with time series $\uu(t)$
 derived from a logistic map to illustrate the properties of
 SFA.  The underlying driving force will always be denoted by $\p$ and may vary
 between $-1$ and $1$ smoothly and considerably slower (as defined by the variance of its time derivative
 (\ref{eq:slowness})) than the  time series $\uu(t)$. The approach follows closely the work of Wiskott~\cite{Wis2003c}, but with more systematic variations in the driving force. 
 
 We consider here a driving force which is made up of two frequency components 
\begin{equation}
	\p_(t) = \frac{1}{2}( \underbrace{\sin(0.0005\nu_f t)}_{=\pslow(t)}  + 
												\underbrace{\sin(0.0047\nu_f t)}_{=\pfast(t)}		) \quad\in\quad [-1, 1]
	\label{eq:gamma}
\end{equation} 
where the first component $\pslow$ is roughly ten times slower than $\pfast$.
The question is whether SFA as the driving force detector will detect solely the slower component $\pslow$ of the driving force (in an attempt to minimize $\eta$) or the full driving force $\gamma$ (in an attempt to extract the underlying system dynamics as accurate as possible). A second question is whether a phase transition between both choices might occur as we vary the base frequency $\nu_f$. 

In order to inspect visually the agreement between a slow SFA-signal and the driving force $\p$ we must bring the SFA-signal into alignment with $\p$ (the scale and offset of slow signals $y(t)$ extracted by SFA are arbitrarily fixed by the constraints and the sign is random). Therefore we define an {\em $\p$-aligned signal}
\begin{equation}
	A(y(t);\p) = ay(t)+b
	\label{eq:align1}
\end{equation}
where the constants $a$ and $b$ are chosen in such a way that 
\begin{equation}
	\left\langle (ay(t)+b - \p(t))^2 \right\rangle \stackrel{!}{=} Min.
	\label{eq:align2}
\end{equation}

\subsection{Logistic map in chaotic regime}

\begin{figure*}[htbp]

\centerline{\includegraphics[width=0.97\textwidth]{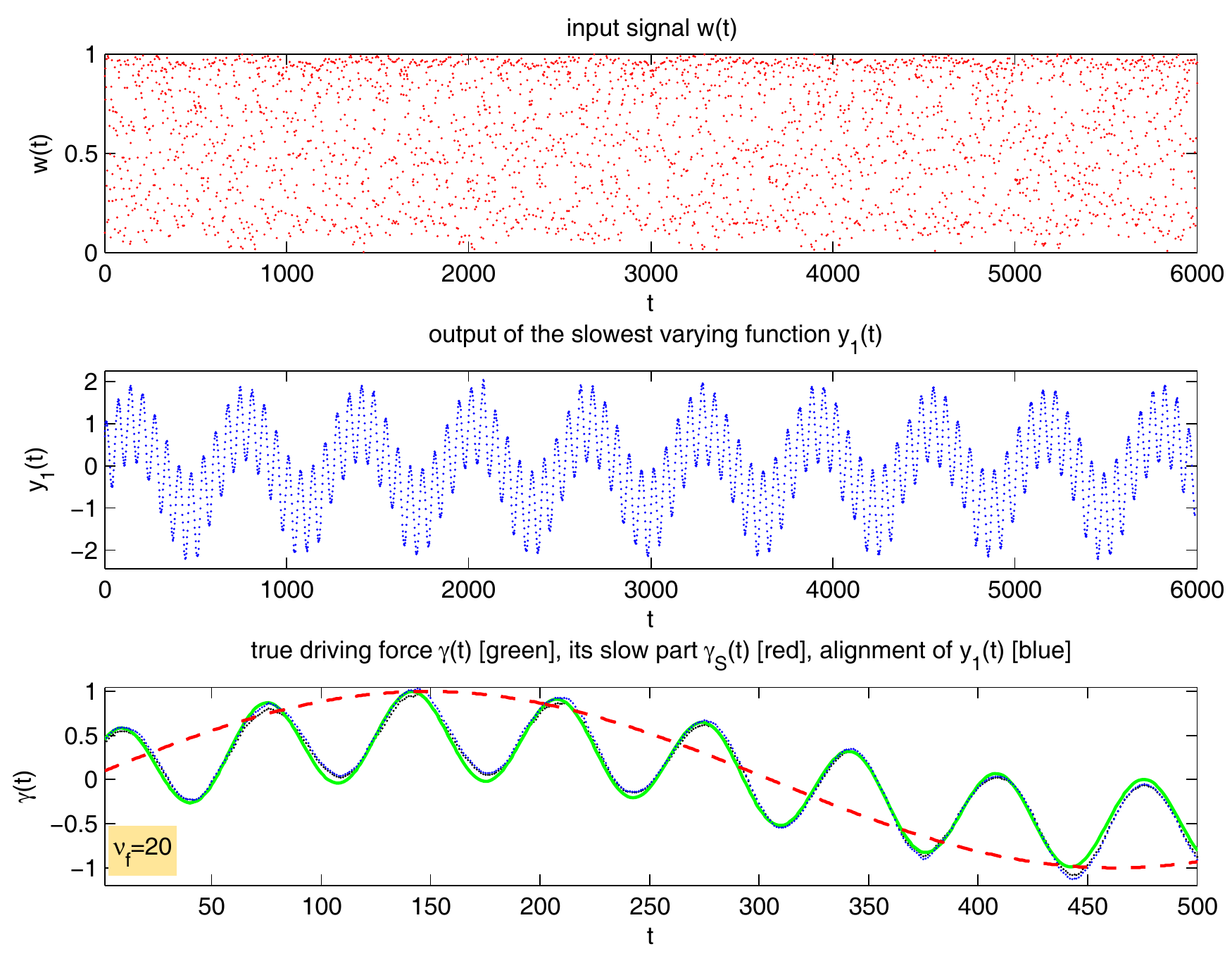}}
%%%%
%%%% generated with [cor,slowness,eta] = drive1(19,1,20,0.1,2) 
%%%% then 'print -f1 -depsc2 -r300 images/logmap-drive-m19-nuf20.eps' and convert with Adobe Distiller
%%%%

 \captionEM{\label{fig:SmoothLogistic} Time series $\uu(t)$ (top) derived from the logistic map with the driving force for $\nu_f=20$, the slowest SFA signal $y_1(t)$ (middle, blue dots), a detail plot for $0<t<500$ of the true slowly varying driving force  $\p(t)$ (bottom, solid green line), its slower component $\pslow(t)$ (bottom, dashed red line)  and the alignment (see \eqref{eq:align1}, \ref{eq:align2}) of $y_1(t)$ to both components (bottom, blue dots). The $\p$-aligned output $A(y_1(t);\p)$ ( = estimated driving force $\pest(t)$) is here in good agreement with the green component. The absolute correlation  between $y_1(t)$ and the true driving force is $\left|C(y_1(t),\p(t))\right|=0.9961$. SFA was done with embedding dimension $m=19$.}
\end{figure*}

 We  consider a time series derived from a logistic map
\begin{equation}
  \uu(t+1) = (3.9+0.1 \p(t)) \uu(t) (1-\uu(t)) \,,
  \label{eq:logistic1}
\end{equation}
 which maps the interval $ [0,1] $ onto itself and has the shape of an upside-down parabola crossing the
 abscissa at 0 and 1.  
Parameter $r(t) = 3.9+0.1 \p(t) $ governs the height of the parabola. Since the logistic map is in its chaotic regime for $r>3.57$ we have with $r(t)  \in [3.8,4.0] $ a highly chaotic map with no visible structre (Figure~\ref{fig:SmoothLogistic}, top).
 
 Taking the time series $\uu(t)$ directly as an input signal would not give
 SFA enough information to estimate the driving force, because SFA
 considers only data (and its derivative) from one time point at a time.
 Thus it is necessary to generate an embedding-vector time series as an
 input.  Here embedding vectors at time point $t$ with delay $\tau$ and
 dimension $m$ are defined by
\begin{eqnarray}
 \vec{\emb}(t) &:=& [	\uu(t - \tau (m-1)/2),\, 
 											\uu(t - \tau ((m-1)/2-1)),\, ..., 
 											\uu(t + \tau (m-1)/2)]^T \,,
\end{eqnarray}
 for scalar $\uu(t)$ and odd $m$.  The definition can be easily extended to
 even $m$, which requires an extra shift of the indices by $\tau/2$ or its
 next lower integer to center the used data points at $t$.  Centering the
 embedding vectors results in an optimal temporal alignment between
 estimated and true driving force.

 The following simulations are based on 6000 data points each and were done
 with \textsc{Matlab 7.0.1} (Release 14) using the SFA toolkit {\tt sfa-tk}~\cite{Berkes2003}.
 
\begin{figure*}[htbp]

\centerline{\includegraphics[width=0.97\textwidth]{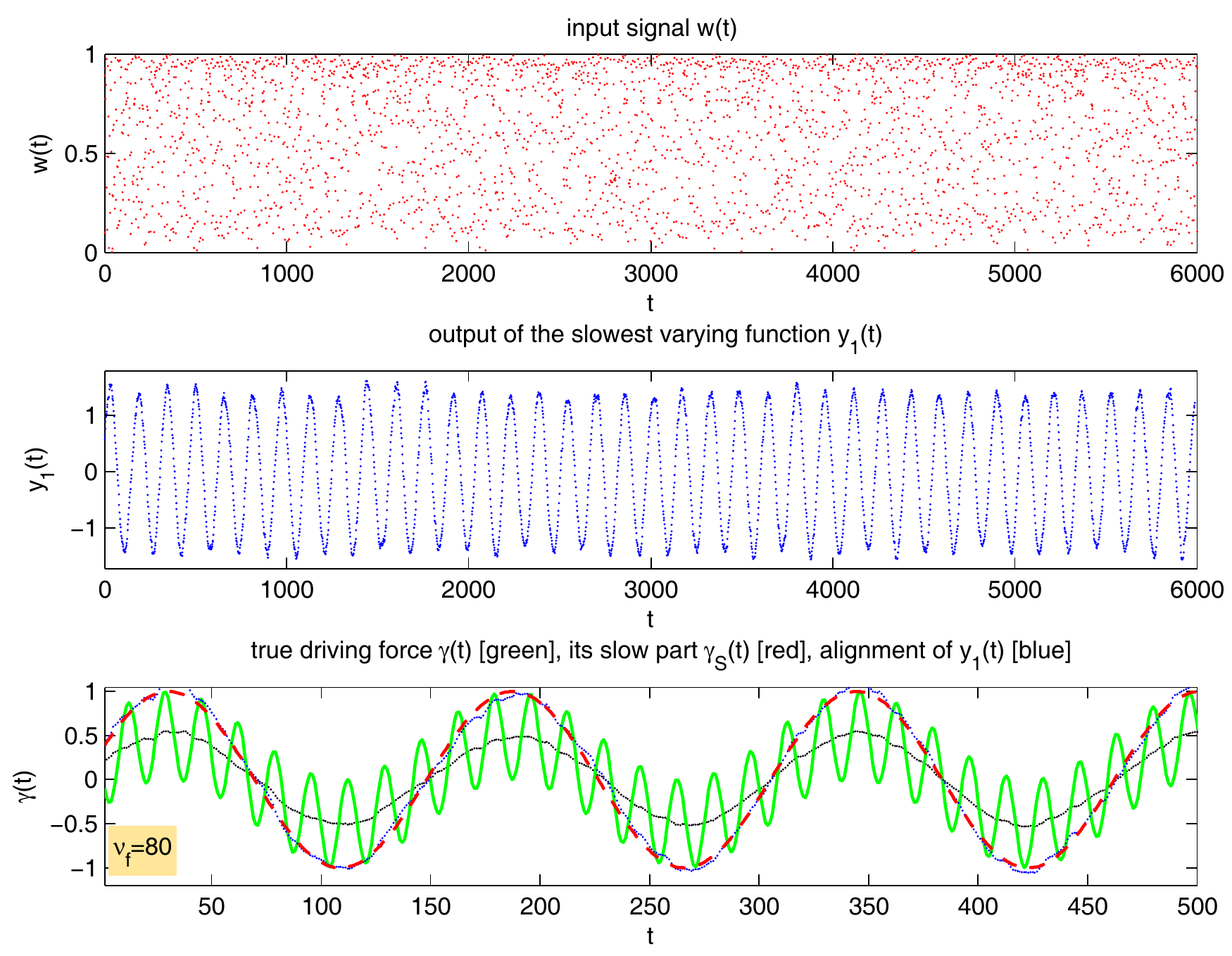}}
%%%%
%%%% generated with [cor,slowness,eta] = drive1(19,1,80,0.1,2)
%%%% then 'print -f1 -depsc2 -r300 images/logmap-drive-m19-nuf80.eps' and convert with Adobe Distiller

 \captionEM{\label{fig:SmoothLogisticFast} Same as Fig.~\ref{fig:SmoothLogistic} but with base frequency $\nu_f=80$. The estimated driving force (bottom, blue dots) is now in good agreement with the slower red component $\pslow(t)$. (We see two blue dotted curves since we align the slowest SFA signal once with $\p(t)$ and once with $\pslow(t)$.) The absolute correlation  between $y_1(t)$ and the slow component is $\left|C(y_1(t),\pslow(t))\right|=0.9974$.}
\end{figure*}

 Figure~\ref{fig:SmoothLogistic} shows the time series, the estimated driving force (from SFA with $m=19$, $\tau=1$ and second order monomials), and the true driving force. The estimated driving force is at $\nu_f=20$ in good alignment with the true driving force. If we use a four times higher base frequency $\nu_f=80$ in Figure~\ref{fig:SmoothLogisticFast} we see that the estimated driving force is now in nearly perfect alignment with the slower component $\p_S(t)$. This is remarkable since the slower component is not directly visible in the driving force, only indirectly as envelope around the green curve (see detail plot in Figure~\ref{fig:SmoothLogisticFast}, bottom).
 
 \subsection{Logistic map in predictable regime}
\begin{figure*}[htbp]

\centerline{\includegraphics[width=0.97\textwidth]{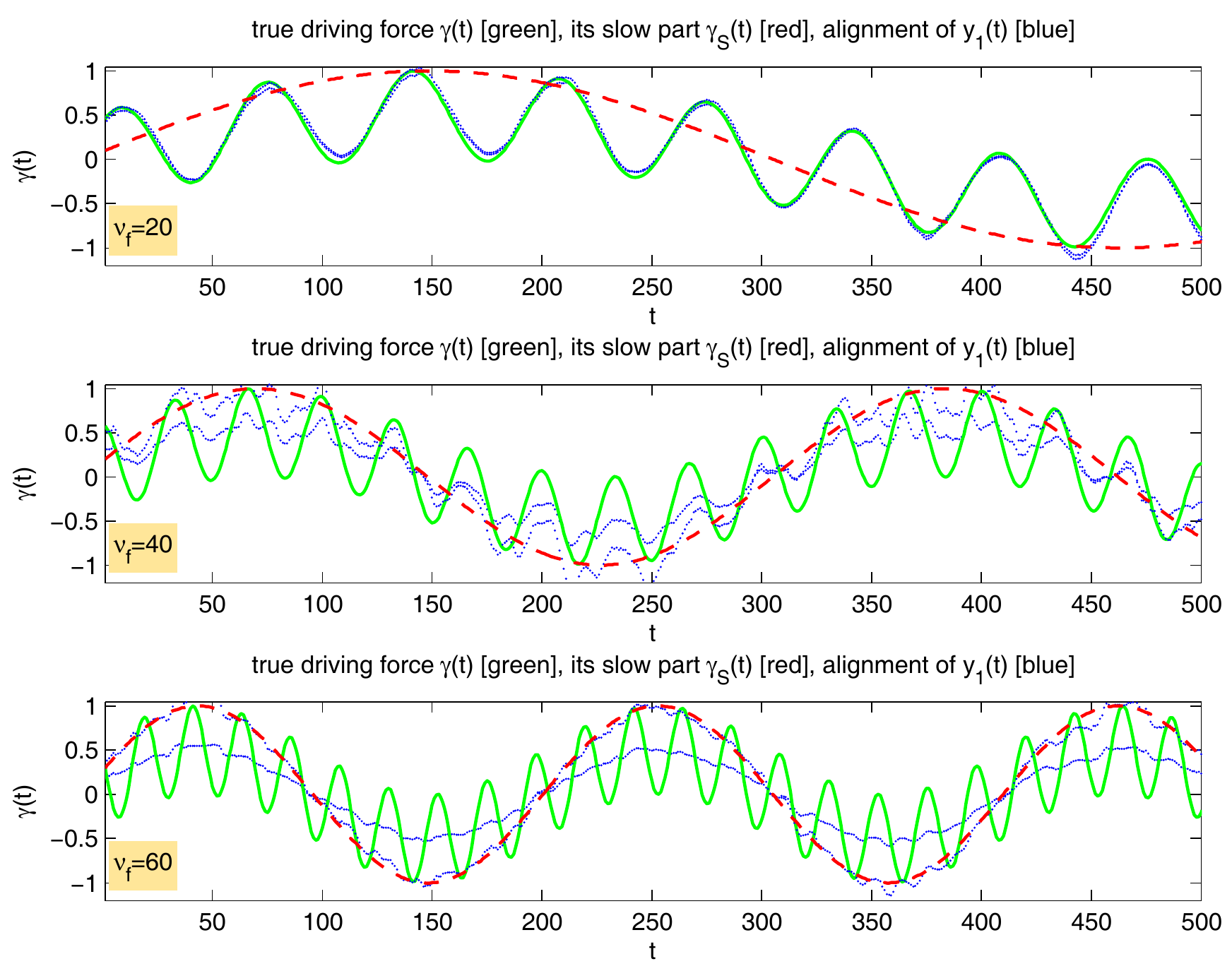}}
%%%%
%%%% Fig. 3
%%%% generated with drive1_sweep(19,0.1,20,80,20,2); then 
%%%% 'print -f4 -depsc2 -r300 images/logmap-drive-m19-nuf20-40-60.eps' and convert with Adobe Distiller
%%%%

 \captionEM{\label{fig:LogisticPhaseTrans} The bottom panel of Fig.~\ref{fig:SmoothLogistic} for three  base frequencies $\nu_f=20, 40, 60$ clearly shows the phase transition from the complete driving force $\p(t)$ (green solid line) to its slower subcomponent $\pslow(t)$ (red dashed line). (We see two blue dotted curves since we align the slowest SFA signal once with $\p(t)$ and once with $\pslow(t)$.)}
\end{figure*}

\begin{figure*}[htbp]
	
\centerline{\includegraphics[width=0.47\textwidth]{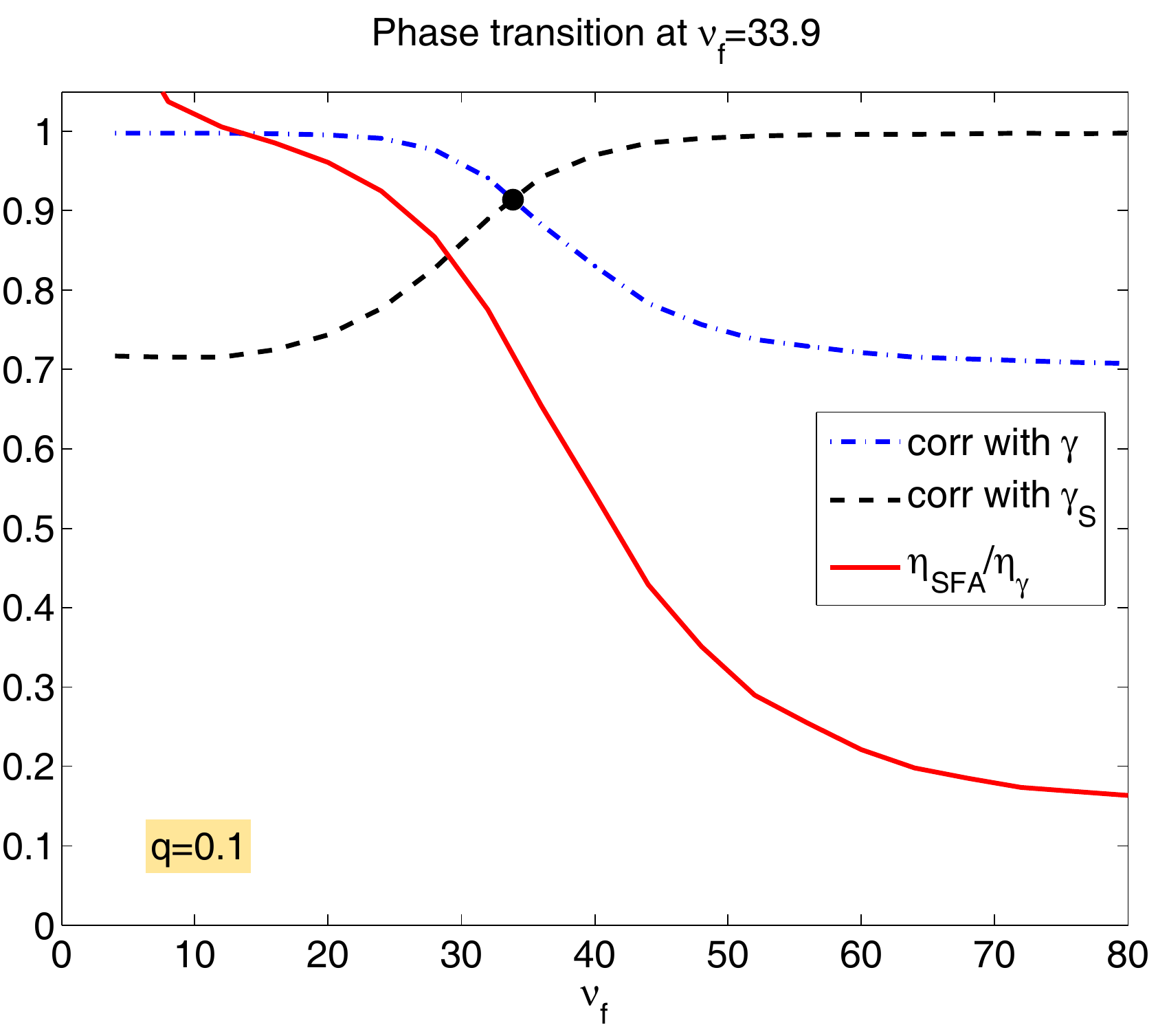}
						\includegraphics[width=0.47\textwidth]{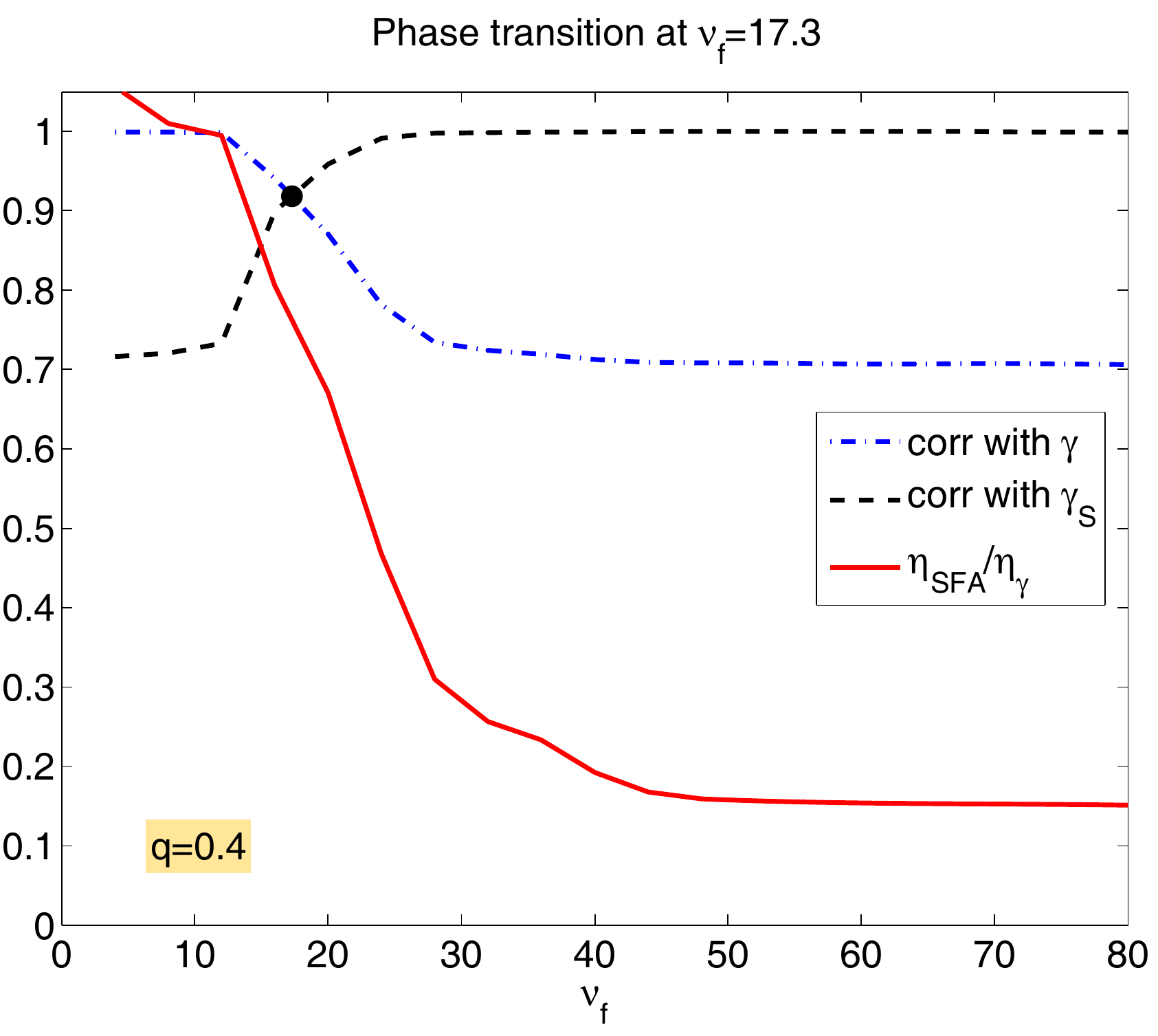}}
%%%% Fig 4, 
%%%% generated with drive2_sweep, which calls drive1_sweep.m with m=19, fmax=80, fstep=4, 
%%%% fmin=fstep, tau=1 and with q=0.1 or q=0.4, resp., in the w-defining command;
%%%% then 'print -f2 -depsc2 -r300 images/logmap-phasetrans-m19-q01[-q04].eps' 
%%%% and convert with Adobe Distiller
%%%%
	
	\captionEM{\label{fig:phasetrans} Absolute correlation $\left|C(y_1(t),\p(t))\right|$ and $\left|C(y_1(t),\pslow(t))\right|$ of the slowest SFA-signal $y_1$ with the driving force $\p$ and with its slow component $\pslow$ (dash-dotted and dashed lines). For sufficient large $\nu_f$ the correlation with the slow component $\pslow$ is stronger than the correlation with $\p$. Left: $q=0.1$ (logistic map in its chaotic regime). The quotient $\eta_{SFA}/\eta_\p = \eta(y_1)/\eta(\p)$ (solid red line), being initially 1 or higher, drops largely for $\nu_f>36$ and eventually approaches a value of only $0.1$, which corresponds to $\eta(\pslow)/\eta(\p)$. SFA was carried out with monomials of degree~2, embedding dimension $m=19$ and delay $\tau=1$. Right: The same for $q=0.4$ (logistic map, only partly chaotic) gives the same qualitative results, only the phase transition is shifted to lower frequency $\nu_f=17$.}
\end{figure*}

 Fig.~\ref{fig:LogisticPhaseTrans} shows quite clearly that there is a phase transition occuring around $\nu_f=40$. 
 %A short movie showing the development of such a phase transition as $\nu_f$ is varied is available electronically as accompanying material~\cite{Kon09-MovieArXiV}.  %%%% TODO!
 To localize the phase transition and to study its  dependence on other parameters of the system we first generalize the logistic map \eqref{eq:logistic1} by a parameter $q \in [0.1,3.9]$ allowing to control the presence or absence of chaotic motion: 
\begin{equation}
  \uu(t+1) = (4.0-q+0.1 \p(t)) \uu(t) (1-\uu(t)) \,,
  \label{eq:logistic2}
\end{equation}
which reduces to \eqref{eq:logistic1} for $q=0.1$. Parameter $q$ controls the predictabilty of the logistic map:
For $q<0.33$ the map is fully in its chaotic regime, for $0.33<q<0.53$ we have a mixture of chaotic and  predictable periods and for $0.53<q<3.9$ it is long-term predictable. In Fig.~\ref{fig:phasetrans} we vary the base frequency $\nu_f \in [20,80]$ and we define the phase transition frequency $\nu(P.T.)$ as the lowest $\nu_f$ with 
$\left|C(y_1(t),\p(t))\right| < \left|C(y_1(t),\pslow(t))\right|$, where $C(\cdot,\cdot)$ denotes the usual correlation. Fig.~\ref{fig:phasetrans}, right, shows that for small $q=0.1$ (fully chaotic $\uu$) a a phase transition occurs at $\nu_f=36$ while for larger $q=0.4$ (mix of chaotic and non-chaotic periods in $\uu$) the phase transition happens earlier at $\nu_f=20$ (Fig.~\ref{fig:phasetrans}, left).   

\subsection{The phase transition as a function of q and m}

\begin{figure*}[htbp]
\centerline{\includegraphics[width=0.47\textwidth]{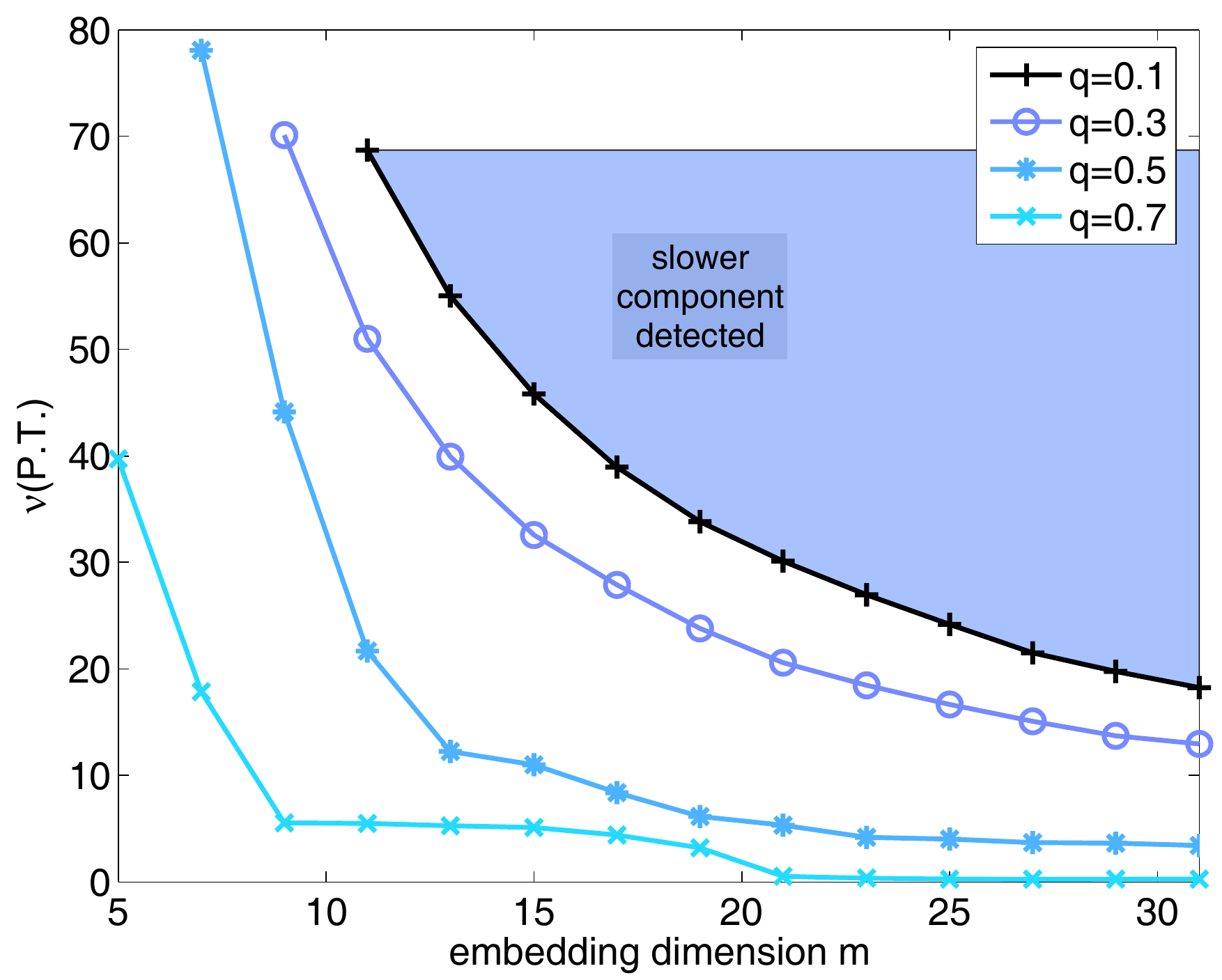}
						\includegraphics[width=0.47\textwidth]{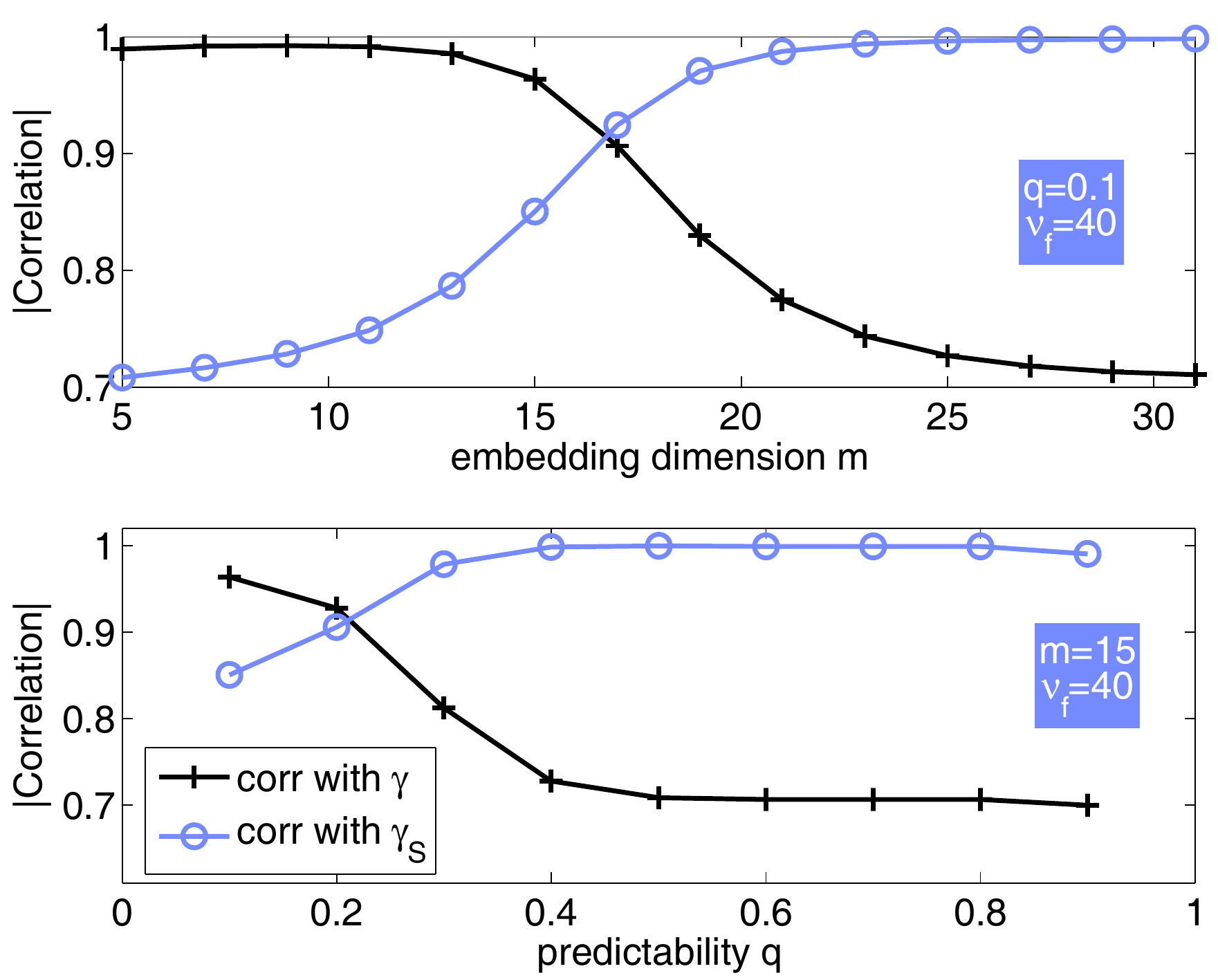}}
%%%%
%%%% generated with multi1[2]_sweep, which calls drive1_sweep.m for many values of m and q
%%%% then 'print -f1 -depsc2 -r300 images/phasetrans_multi1[2].eps' and convert with Adobe Distiller
%%%%
	
	\captionEM{\label{fig:PT_multi} Left: Phase transition frequency $\nu(P.T.)$ as a function of $q$ and $m$. In the area above each phase transition curve the detection of the slower component $\pslow$ of the driving force is stronger. Right: The absolute correlation values at fixed $\nu_f=40$ show a phase transition towards the slower component as we increase the embedding dimension $m$ (top) or the value $q$ ('predictability' of the logistic map, bottom).}
\end{figure*}

How does the 
phase transition frequency $\nu(P.T.)$ vary as a function of the predictability $q$ and the embedding dimension $m$ of the SFA-input signal? Both parameters are varied systematically over a broad range and the results are depicted in Fig.~\ref{fig:PT_multi}. First of all it is interesting to note that the SFA algorithm, being basically parameter-free, works very successfully over this broadly varying input material, which makes SFA a robust and versatile algorithm. Tab.~\ref{tab:eta} shows the slowness indicator $\eta$ for some $m$ and $q$.

Before we discuss the results further in \Secref{sec:discuss}, a second remark is in order concerning the SFA implementation {\tt sfa-tk}~\cite{Berkes2003}: While it worked well for small embedding dimensions $m$, larger $m$ led quite inevitably to wrong "slow" signals $y_1$ which were neither slow nor did they respect the unit variance condition $\langle\os^2\rangle=1$. In an accompanying second paper´~\cite{Kon09b} we trace this behaviour back to numeric instabilities of the implementation and present a slightly modified implementation (closer along the lines of~\cite{WisSej2002}) and based on SVD which  successfully avoids these numeric instabilities. This modified implementation is used throughout the experiments in this paper. 

\section{Discussion}
\label{sec:discuss}
It is important for driving force analysis with SFA to understand the mechanisms by which the slowest signal is selected. If the driving force contains two components of different frequencies two interesting things might happen: If the base frequency $\nu_f$ is large enough then SFA will return the slower component as slowest signal. This is quite remarkable, since SFA detects a signal with a smaller $\eta$ than the driving force itself. Recall that this slower component is not directly visible in the driving force, only indirectly as the envelope. But after all, it is also quite understandable: If we view the dynamical system as a two-stage process where the slow component $\p_S$ is considered as a modulating force acting on the other (faster) component $\p_F$ with the output of this stage acting on the dynamical system, then in such a system description the slower component $\p_S$ becomes directly visible. 

Surprisingly, if we lower the base frequency $\nu_f$, we reach the point where the slow component comes "out of sight" and the slowest signal returned by SFA is well-aligned with the driving force itself (slow plus fast component). Why is the slow component alone no longer detected by SFA? We hypothesize that two reasons are responsible for this: 
\begin{itemize}
	\item The embedding dimension $m$ might be too low so that the slow component is perceived within the embedding vector as nearly constant. Then increasing $m$ should lower the phase transition frequency and finally detect the slow component. 
	\item Another reason might be the chaotic nature of the logistic map. The slow component might vary on a time scale which is slower than the time scale of 'forgetting' in the chaotic logistic map and thus this component alone is not detectable by SFA. If this is true then moving to a better predictable region of the logistic map (increasing $q$) should make the slow component again detectable. 
\end{itemize}
Both hypotheses are well-supported by the result shown in Fig.~\ref{fig:PT_multi}. On the left-hand side we see the location of the phase transition. For most input signals which are a function of $q$ and $\nu_f$ there seems to be a sufficient large $m$ so that the slow component becomes detectable. For $q=0.7$ this occurs already at very low frequencies. The curve for $q=0.6$ (not shown) is for $m>10$ very similar to $q=0.7$, which is well-understandable if we recall that all $q>0.53$ make the time series long-term predictable, thus even a very slow subcomponent becomes detectable. On the right-hand side of Fig.~\ref{fig:PT_multi} we see that both methods, increasing $m$ or increasing $q$ finally lead to a reliable detection of the slow subcomponent as it is claimed by our hypothesis.

\paragraph{Other slow signals}

Consider the case where the slowest signal $y_1$ aligns well with the slow component $\p_S$. One might expect that the second slow signal $y_2$ aligns well with the fast component $\p_F$. But it turns out, that usually the next slow signals $y_2, y_3, y_4, \ldots$ have a similar slowness as $\p_S$ although they are orthogonal to $y_1$. Only at some later index, e.g. $y_5$ or $y_{10}$, corresponding to the 5th or 10th smallest eigenvalue, the slowness suddenly jumps to higher values and reveals a signal well-aligned with $\p_F$. Note that if the slowest signal $y_1$ has a high correlation with $\p_S$, no other signal will align perfectly with the complete driving force because this would be not orthogonal to $\p_S$ and is thus avoided.

%Note: the number of SFA-outputs varies greately: from 107 for q=0.5 to 209 (!!) for q=0.3 (both with m=19). Indeed 209 is the maximum dimension from the 2nd-degree monomial formula X_p = m + m(m+1)/2 (see notes_SFA.doc). The more we go to the predictable regime the more EV are discarded in SVD. (It can be traced back to zeros from index 108..209 in  Dmtx (lcov_pca2.m) and correspondingly reduction to 107 dimensions in sfa2_step.m

\paragraph{Robustness of SFA}

 SFA as tested in this paper works robustly over a large range of parameters $\nu_f, m, q$. It is however necessary to deal carefully with zero eigenvalues which occur frequently when the embedding dimension $m$ is large and/or the noise is low. If zero eigenvalues are not handled it is likely to see numerical instabilities. A numerically robust way to handle them is based on SVD and is described in~\cite{Kon09b}.
  
\begin{table}[t]
	\captionEM{ Slowness indicator $\eta$ for various $m$ (embedding dimension) and $q$ (predictability). Large $\eta$ indicate fast, small $\eta$ indicate slow signals. The true driving force $\p$ has $\eta=127$ while its slow component $\pslow$ has $\eta=19.1$.
	\label{tab:eta}}
\begin{center}
	\begin{tabular}{|r|r|r|r|r|r|r|}
	\hline
	& & \multicolumn{5}{c|}{$\eta$}						\\ \hline
$m$&$q$& 0.1&    0.3&    0.5&    0.6&		 0.7\\ \hline\hline
 5& & 147.85&	133.67&	124.88&	 77.59&	106.44\\ \hline
10& & 123.84&	120.43&	 59.16&	 19.56&	 19.47\\ \hline
15& & 105.63&	 67.35&	 21.27&	 19.29&	 19.08\\ \hline
20& &  59.93&	 30.32&	 19.43&	 18.87&	 19.06\\ \hline
30& &  22.66&	 19.63&	 19.10&	 13.82&	 19.08\\ \hline												   
	\end{tabular}	
	%%%%
	%%%% generated as array eta from sfa-tk/demo/multi3_sweep.m using tau=1;nuf=40.
	%%%% copy array 'eta' from Array Editor and paste it here 
	%%%%
\end{center}
\end{table}

\paragraph{Accuracy of the Estimated Driving Force}

The driving force is estimated with high accuracy, although the estimation is undetermined up to any invertible transformation~\cite{Wis2003c}. We found that his is true even for large embedding dimension, e. g. $m=51$, in contrast to the hypothesis in~\cite{Wis2003c} that only small embedding dimensions would avoid more complicated invertible transformations.

\paragraph{Noise Sensitivity}

The results described in this paper were obtained with noise-free data. We tested in some simulations the effects of adding Gauusian noise to the data before embedding. For $m=19, \nu_f=56$ and $q=0.4$ the effect of adding 1\%, 2\% and 5\% noise brought the correlation between slow component $\p_S$ and slowest SFA-signal $y_1$ from $|C|= 0.999$ down to 0.85, 0.75 and 0.60, resp. Thus for small noise levels $<1\%$ the main effects persist, but in contrast to~\cite{Wis2003c} the noise sensitivity is somewhat higher, which means that 5\% noise destroy most of the correlation with the slow component. This might be due to the more complex nature of the driving force build up from multiple components. 
On the other hand we found that larger embedding dimensions, e. g. $m=51$ stabilize the system and bring up the correlation again to 0.963, 0.893 and 0.804, resp.,
%  1%     2%     5%
%  0.963  0.898  0.804
but further experiments are needed to investigate this systematically.

\paragraph{High-Dimensional Input Data}

As the preceeding paragraph has shown higher dimensional input data usually improve the SFA performance. However, since the number of monomials grows quadratically with the embedding dimension, the requirements in computer memory and computing time quickly increases. It is therefore interesting to investigate whether similar results as with high $m$ can be obtained by hierarchical approaches where first smaller parts of the embedding vector are analyzed and then combined by a final SFA, as has been demonstrated in~\cite[]{WisSej2002}.

\paragraph{Connection to human perception}

Since SFA has been originally developed as a model for neural information processing~\cite[]{Wis98a}, it might be natural to ask, whether the observed switch between components and its phase transition has any parallel in human perception or human motion coordination. Several phenomena with switching effects are discussed in the literature: 
\begin{itemize}
	\item The well-known backward spinning-wheel illusion~\cite{PurPayAnd1996} occurs frequently in movies or under stroboscopic lighting conditions and it shows the transition from a fast forward rotation percept to a slow backward rotation percept. This effect is usually explained by the snapshot-like presentation of the percept which has ambiguous motion interpretations. Somewhat less known is that a similar, although harder to perceive effect can occur under plain sunlight and direct view with the eye~\cite{KliHolEag2004,PurPayAnd1996}. No snapshot-like explanation is possible here, the percept is continous having a greater resemblance to the smoothly varying driving force of our SFA experiments. A  possible explanation of the sunlight spinning-wheel illusion is according to~\cite{KliHolEag2004} that rivalry between different motion detectors in the brain occurs.
	\item Another well-known phase transition occurs in bimanual motion coordination when performing certain movements with the index fingers of both hands~\cite{Kelso1981} for which a theoretical model exists, the so-called Haken-Kelso-Bunz model~\cite{HKB1985}. The Haken-Kelso-Bunz model also describes a phase transition and certain hysteresis effects. 
\end{itemize}
SFA has shown similar capabilities in the sense that the same setup can learn to synchronize with different components of a driving force, depending on the experiment conditions. It remains however to be studied, whether {\em one} trained SFA system can (without further learning) switch between different components when applied to signals with smoothly varying base frequency and whether a hysteresis effect can be observed.

\section{Conclusion}

 In this paper we have investigated the notion of {\em slowness} in  SFA. It has 
been verified that SFA can reliably detect either slow driving forces or their subcomponents over a broad range of parameters in nonstationary time series, even in the presence of chaotic motion. 
 
 However it has also been seen that what is perceived as {\em slow} can vary for driving forces made up of components on different time scales: Depending on the embedding dimensions and the predictability of the underlying dynamical system we observe phase transitions where the slowest SFA-signal moves from alignment to a slow subcomponent to alignment with the (faster varying) complete driving force. Notably, when alignment to the slow subcomponent occurs, SFA is capable of detecting slow signals with an $\eta$-indicator considerably lower than the $\eta$-value of the true driving force.
 
 There are still a number of open questions.  
 Does hierarchical SFA, which achieves larger embedding dimensions with a smaller computing budget, allow for the detection of very slow components which are 'out-of-reach' for plain SFA? Can an extended SFA system model more closely certain switching effects known from human perception (as for example the backward spinning-wheel illusion~\cite{PurPayAnd1996})? Finally, it is necessary to apply SFA to real world data and to see whether the results reported in this study have some similar parallel there. 
 
 In real world data it will often not be possible to vary the base frequency or the degree of nonlinearity in the observed dynamical system systematically. Therefore, one advice from the present study should be to vary the embedding dimension over a broad range in order to detect possible slow signals which otherwise might be hidden.
In any case, SFA has shown to be robustly working on a broad range of input data and it is able to reveal subtle components in the driving forces, thus making it a versatile tool for driving force detection.   

%TODO
%\begin{itemize}
%	\item Video of phase transition?
%\end{itemize}

\section{Acknowledgment}

 We are grateful to Laurenz Wiskott for helpful discussions on SFA and to Pietro Berkes for
 providing the \textsc{Matlab} code for {\tt sfa-tk}~\cite{Berkes2003}.  This work
 has been supported by the Bundesministerium f\"ur Forschung und Bildung (BMBF) under the grant 
 SOMA (AIF FKZ 17N1009, "Systematic Optimization of Models in IT and Automation") and by the Cologne University of Applied Sciences under the research focus grant COSA.

% --- this is the normal way: run BibTex on SFA.bib to produce main.bbl:
%\bibliographystyle{alpha}
%\bibliography{SFA}

% --- bititems.tex is a copy of main.bbl, which is needed for arXiv.org (which does not run BibTex)

\end{document}